\pdfoutput=1
\documentclass[11pt]{article}

\usepackage{acl}

\usepackage{times}
\usepackage{latexsym}
\usepackage[T1]{fontenc}
\usepackage[ruled,vlined]{algorithm2e}

\usepackage{tabularx}
\usepackage{booktabs}
\usepackage[utf8]{inputenc}
\usepackage{microtype}
\usepackage{inconsolata}
\usepackage{graphicx}
\usepackage{multirow}
\usepackage{mdwlist}
\usepackage{capt-of}

\newcommand{\emoclass}{\texttt{Emotion}\xspace}
\newcommand{\std}[1]{{\fontsize{8pt}{36pt}\selectfont({#1})}}

\title{LLM-based Affective Text Generation Quality\\ Based on Different Quantization Values}

\author{Yarik Menchaca Resendiz$^{1,2}$ \and Roman Klinger$^{2}$ \\
  $^{1}$Institut f\"ur Maschinelle Sprachverarbeitung, University of Stuttgart, Germany\\
  $^{2}$Fundamentals of Natural Language Processing, University of Bamberg, Germany\\
  \texttt{yarik.menchaca-resendiz@ims.uni-stuttgart.de,roman.klinger@uni-bamberg.de}
  }

\begin{document}
\maketitle
\begin{abstract}

Large language models exhibit a remarkable capacity in language generation and comprehension. These advances enable AI systems to produce more human-like and emotionally engaging text. However, these models rely on a large number of parameters, requiring significant computational resources for training and inference. In some scenarios, accessing these resources can be challenging (e.g., budget or hardware limitations). Techniques like reducing precision bits can make models more memory-efficient, reducing the computational resources needed, at the cost of reduced accuracy. This paper addresses
  the trade-off between different quantization values, GPU RAM
  utilization, and text quality in affective text generation (e.g.,
  ``I really enjoy running in the snow-covered forest''). To evaluate,
  we use an emotion classifier and ten seed prompts to generate
  affective text. We test three setups of precision bits (8, 16, and
  32) across five open-weight language models from two different families. Our findings demonstrate
  that bit reductions lead to memory savings, achieving a reduction
  of 76 \%. However, this optimization comes with a trade-off, leading
  to a decrease of up to 10 pp in F$_1$ score for larger models and an
  increase of 10 pp for smaller models, along with roughly double the
  inference time. In terms of text quality, larger models at lower
  quantization levels generally outperform smaller, higher-precision
  models -- while requiring similar memory.
\end{abstract}

\section{Introduction}

Large Language Models \citep[LLMs,][]{2023llama,
  radford2018improving,NEURIPS2020_1457c0d6,radford2019language} have demonstrated unprecedented capabilities in language generation and contextual comprehension. These advancements have improved the ability of AI systems (e.g.,  virtual assistants,  customer service bots), to generate human-like and emotionally engaging text.
  This ability to produce affective text is crucial for making AI systems capable of building user trust and satisfaction.
   However, the substantial computational resources required by these models often make them inaccessible to individuals or organizations with limited resources.
   A democratizing solution is quantization, which involves converting high-precision values to lower-precision ones (e.g., from 32 to 8 bits) to reduce overall model size \citep{hu2022lora, QLoRA}. This approach makes LLMs more memory-efficient, requiring less GPU memory, storage, and energy. However, the primary drawback of quantization is the potential loss of accuracy, with higher bit reductions posing a greater risk. This reduction in precision sets the stage for exploring the trade-off between computational resources, such as GPU RAM, inference time, and text quality. Our research aims to answer: \textbf{What is the trade-off between precision in bits on LLMs in affective text generation?}

 Post-training quantization is popular due to its simplicity and
 effectiveness. \citet{Jacob2018} demonstrated that it can reduce full-precision models to 8-bit integers with minimal accuracy loss,
 making it suitable for use on edge devices (e.g., smartphones, tablets). Quantization-aware training incorporates quantization during training, producing models more robust to precision reduction. \citet{Courbariaux2015BinaryConnectTD} introduced BinaryConnect, a quantization-aware method that trains neural networks with binary weights, significantly reducing memory usage and computational complexity.

Quantization methods are essential for complex tasks such as emotion-conditioned text generation, which enhances human-computer interaction but demands significant computational resources.  Most research focuses on sentiment polarity \citep{zhang2019emotional, maqsud2015synthetic, niu2018polite}, with fewer studies on emotion conditioning \cite{ghosh-etal-2017-affect, song2019generating, zhou2018emotional, cui2022modeling, colombo-etal-2019-affect, menchaca-resendiz-klinger-2023-affective}.

None of the previous work has focused on the trade-off between affective text generation and quantization, which is crucial for enabling AI assistants to operate on devices with limited resources (e.g.,  smartphones, and laptops). Our
research addresses this gap by investigating the trade-offs between
GPU RAM, inference time, and text quality when using different
precision bits in affective text generation.

\section{Methodology}
\label{sec: method}
The presented work aims to provide an understanding of affective text
generation using multiple precision bits\footnote{The code can be
  found at Blinded for review.}. 
Section \ref{sub:seed} introduces the seed prompts used as affective
generation prompts for the LLMs. Section \ref{sub:
  precision} describes the LLMs and the quantization
configurations. Section \ref{sub: classifier} explains the automatic
evaluation setup. Finally, Section \ref{sub: text quality} presents the text
quality evaluation.

\subsection{Seed Prompts}
\label{sub:seed}
We select a set of 10
seed prompts strategically designed to guide the language models for
effective affective text generation. This collection comprises a
mix of straightforward prompts (e.g., Generate a text that expresses \emoclass, Phrases that express \emoclass) that users can easily conceive, alongside prompts drawn from psychological studies, such as the ISEAR survey (e.g., "Please describe a situation or event—in as much detail as possible—in which a reader felt \emoclass?"). The following diverse combination ensures a comprehensive exploration of emotional expressions, encompassing both intuitive and scientifically validated inputs.

\begin{enumerate*}\footnotesize
    \item Describe a situation where a person felt \emoclass.
    \item Write a text that expresses \emoclass.
    \item Phrases that express \emoclass.
    \item What is a sentence example for \emoclass?
    \item Can you provide an example of a situation where someone experienced \emoclass?
    \item What is an example of a \emoclass sentence?
    \item \emoclass sentence.
    \item Experience for \emoclass?
    \item Please describe a situation or event—in as much detail as possible—in which a reader felt \emoclass.
    \item Please complete the sentence: I felt \emoclass when/because.
\end{enumerate*}

\subsection{Precision Bits}
\label{sub: precision}
To address the research question, \textbf{What is the trade-off
  between precision in bits on LLMs in affective text generation?}, we
utilize Llama 2 Chat \citep{2023llama} with three parameter sizes (7B,
13B, and 70B) and Mistral 7B \cite{jiang2023mistral} and Mixtral-8x7B
\citep{jiang2024mixtral}. Each model is loaded into memory using three
different precisions: 8, 16, and 32 bits -- the default setting. We
employ a total of 15 configurations to generate affective text from
the seed prompts.

\subsection{Text Generation}

The text is generated using the LLMs and the 10 seed prompts. We
design an \textit{Emotion-Prompt} that integrates the seed prompt (for
instance, ``Write a text that expresses \emoclass.'') with one of
seven specified emotions: Joy, Fear, Anger, Sadness, Disgust, Shame,
and Guilt. In this setup, \emoclass is replaced by each emotion
category (see Table \ref{tab:emotion_prompts} in appendix \ref{appendix:affective text generation}, for examples). For every
\textit{Emotion-Prompt}, we employ a beam search strategy to generate the three most likely sentences -- 210 sentences in total. This process uses a beam size of 30, sets the next-token temperature to 0.7, and applies a top-p (nucleus) sampling strategy at 0.7. Additionally, we take measures to ensure the generated output is free from sentences that contain duplicate instances of any bigram. The inference tasks for the LLMs models are performed and evenly distributed across a GPU cluster equipped with 8 NVIDIA RTX A6000 GPUs, for 10 times per model configuration.

\subsection{Automatic Classification}
\label{sub: classifier}
To evaluate the performance of the LLMs configuration automatically,
we use an emotion classifier fine-tuned on the ISEAR dataset \cite{scherer1994evidence}.
The emotion classifier is built on top of RoBERTa
\citep{liu2019roberta} using the standard parameters for ten epochs
with a batch size of five. Please refer to Table~\ref{tab:emo_scores}
for precision, recall, and F$_1$ scores. Despite the limited data, these scores are comparable to previous experiments
\cite{Troiano2023}.

\begin{table}
  \centering\small
  \setlength{\tabcolsep}{8pt}
  \renewcommand{\arraystretch}{0.8}
  \begin{tabularx}{\linewidth}{Xrrr}
    \toprule
    \textbf{Emotion} & \textbf{Precision} & \textbf{Recall} & \textbf{F$_1$} \\ 
    \midrule
    Anger & .72 & .58 & .64 \\
    Disgust & .74 & .80 & .77 \\
    Fear & .78 & .93 & .85 \\
    Guilt & .56 & .71 & .62 \\
    Joy & .91 & .92 & .98 \\
    Sadness & .91 & .87 & .89 \\
    Shame & .66 & .43 & .52 \\
    \cmidrule(r){1-1}\cmidrule(rl){2-2}\cmidrule(l){3-3}\cmidrule(l){4-4}
    Macro-Avg. & .75 & .75 & .75 \\ \bottomrule
  \end{tabularx}
  \caption{Precision, Recall and F$_1$ scores from the emotion
    classifier over the 7 emotion classes.}
  \label{tab:emo_scores}
\end{table}

\subsection{Text Quality}
\label{sub: text quality}
We use GPT-4o (see Table  \ref{table:text quality}) and GPT-3.5 (Table \ref{table:gpt 3 text quality} in Appendix \ref{appendix:gpt3 QE}), known for its human-like performance in text quality
evaluation \cite{chiang-lee-2023-closer, liu-etal-2023-g}. This
evaluation focuses on Coherence, Fluency, Grammar, Native Speaker
Likeness, Human Likeness, and Plausibility. To quantify the
evaluation, we adopt a 5-point Likert scale, ranging from 1 (disagree) to 5 (agree), to rate each dimension of text
quality.

\section{Experimental Results}
\label{sec: Experiments}

We investigate the trade-offs involved in using quantization with Large Language Models (LLMs) for affective text generation. Our evaluation includes the trade-offs between quantization and performance (Section \ref{sec: affective text generation}), quantization and text quality (Section \ref{sec: quantization vs text quality}), performance and GPU (Section \ref{sec: f1_vs_gpu}), and GPU and inference time (Section \ref{sec: f1_vs_time}).

\subsection{Quantization vs. F1-score in Affective Text Generation}
\label{sec: affective text generation}

The analysis between quantization levels and F$_1$-scores across
different models shows multiple trends (see Table
\ref{table:quantization performance}). Llama-2-7b's F$_1$-score drops at the highest quantization level (32), while Llama-2-13b shows variability but generally decreases at mid-level quantization (16). In contrast, Llama-2-70b's performance improves with higher quantization, suggesting a more efficient GPU resource usage. Mistral-7B maintains stable high F$_1$-scores across all quantization levels, showing robustness to changes in quantization. Mixtral-8x7B experiences a decline in performance with higher quantization, indicating sensitivity to reduce precision. The analysis suggests that, as expected, quantization reduces the LLM's performance -- by an average of 3.6 pp across all configurations. However, it also results in a 76\% reduction in GPU memory usage.

\begin{table}
\centering
\footnotesize
\begin{tabularx}{\columnwidth}{Xcrrr}
\toprule
\textbf{Model} & \textbf{Qnt.} & \textbf{F$_1$} & \textbf{GPU}& \textbf{I. Time} \std{std}\\
\midrule
Llama-2-7b & 8 & .73  & 7.5 & 20.89 \std{.22}\\
Llama-2-7b & 16 & .73 & 13.4 & 12.53 \std{.82}  \\
Llama-2-7b & 32 & .63 & 26.5 & 15.82 \std{1.8} \\
\midrule
Llama-2-13b & 8 & .73  & 10.5 & 22.92 \std{.73}\\
Llama-2-13b & 16 & .65 & 25.6 & 11.54 \std{.08}\\
Llama-2-13b & 32 & .70 & 42.2 & 16.49 \std{.22} \\
\midrule
Llama-2-70b & 8 & .64 & 50.4 & 34.93 \std{.97}\\
Llama-2-70b & 16 & .72 & 100.5  & 29.72 \std{.49} \\
Llama-2-70b & 32 & .74 & 268.1 & 51.93 \std{.28}\\
\midrule
Mistral-7B & 8 & .73 & 8.9 & 69.99 \std{22.73}\\
Mistral-7B & 16 & .74 & 15.6 & 11.90 \std{.06}    \\
Mistral-7B & 32 & .74 & 29.7 & 33.17 \std{.20}  \\
\midrule
Mixtral-8x7B & 8 & .72 & 46.5 & 125.04 \std{34.03}\\
Mixtral-8x7B & 16 & .66 & 90.9 & 85.89 \std{5.94}\\
Mixtral-8x7B & 32 & .66 & 181.2 & 53.56 \std{.53}\\
\bottomrule
\end{tabularx}
\caption{Performance of Mistral and Llama models at different quantization (Qnt.) values. F$_1$-scores are computed across 7 emotions. GPU usage is in GB. Inference time (I. Time) is in seconds per generated text instance.}
\label{table:quantization performance}
\end{table}

\subsection{Quantization vs. Text Quality}
\label{sec: quantization vs text quality}

Table \ref{table:text quality} illustrates the trade-off between quantization and text quality. Quantization affects text quality differently across models. Llama models generally show a small increase in quality with higher quantization, especially in coherency and plausibility. Mistral-7b remains robust across all quantization levels, while Mixtral-8X7b improves with higher quantization. Larger models at lower quantization values typically outperform smaller models with higher precision -- for example, Llama-2-70b at 8 bits outperforms Llama-2-7b or Llama-2-13b at 16 or 32 bits -- while requiring similar memory.

\begin{table}
\centering
\footnotesize
\setlength{\tabcolsep}{3pt}
\begin{tabularx}{\columnwidth}{Xrrrrrr}
\toprule
\textbf{Model} & \textbf{Qnt.} & \textbf{Flu} & \textbf{N. Spk.} & \textbf{Coh.} & \textbf{Pla.} & \textbf{Gra.} \\
\midrule

Llama-2-7b & 8 & 2.8 & 3.5 & 2.3 & 3.4 & 3.3 \\
Llama-2-7b & 16 & 2.9 & 3.7 & 2.3 & 3.6 & 3.4\\
Llama-2-7b & 32 & 2.7 & 3.6 & 2.2 & 3.7 & 3.4\\
\midrule
Llama-2-13b & 8 & 2.7 & 3.7 & 2.2 & 3.6 & 3.5 \\
Llama-2-13b & 16 & 2.8 & 3.3 & 2.3  & 3.3 & 3.4\\ %
Llama-2-13b & 32 & 2.9 & 3.5 & 2.3 & 3.6 & 3.4 \\

\midrule
Llama-2-70b & 8 & 3.6 & 4.1 & 3.3 & 4.3 & 4 \\
Llama-2-70b & 16 & 3.5 & 4.3 & 3.3  & 4.0 & 3.9\\
Llama-2-70b & 32 & 2.9 & 3.7 & 2.6 & 3.6 & 3.5\\

\midrule
Mistral-7b & 8 & 4.6 & 4.9 &  4.3 & 4.5 & 4.9 \\
Mistral-7b & 16 & 4.6 & 4.8 &  4.3 & 4.6 & 4.8\\
Mistral-7b & 32  & 4.6 & 4.8 & 4.2 & 4.7 & 4.8 \\

\midrule
Mixtral-8X7b & 8 & 4.6 & 4.9 & 4.2 & 4.7 & 4.9 \\
Mixtral-8X7b & 16 & 4.5 & 4.9 & 4.1 & 4.9 & 4.9 \\
Mixtral-8X7b & 32 & 4.5 & 4.9 & 4.2 & 4.8 & 4.9\\

\bottomrule
\end{tabularx}
\caption{Text quality evaluation using a five-level Likert scale for fluency (Flu), native speaker (N. Spk), coherency (Coh.), plausibility (Pla.), and overall grammar (Gra.). A score of 1 indicates "not at all agree," and 5 indicates "extremely agree" (higher is better).}
\label{table:text quality}
\end{table}

\begin{figure}
\centering
\includegraphics[width=\columnwidth]{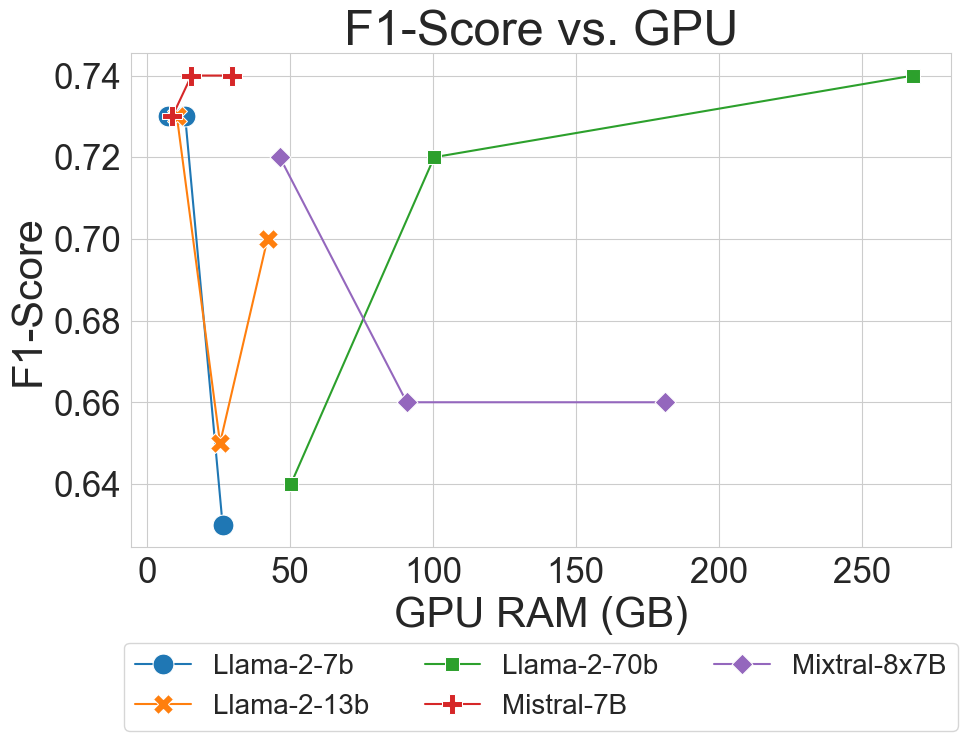}
\caption{F$_1$-score vs. GPU memory usage for different models and
  configurations.}
\label{fig:f1 vs gpu}
\end{figure}

\subsection{GPU Usage vs. F1-score}
\label{sec: f1_vs_gpu}
Figure \ref{fig:f1 vs gpu} compares the performance of Llama and Mistral models across quantization levels (8, 16, and 32 bits) in terms of F$_1$ score and GPU RAM usage. Models with fewer parameters (Llama-2-7b and Mistral-7B) exhibit consistent F$_1$ scores across different quantization values. In contrast, models with more parameters (Llama-2-70b and Mixtral-8x7B) show greater variability in F$_1$ scores with changes in quantization. Generally, reducing model precision can decrease performance by up to 10 percentage points (Llama-2-70b) or increase it by 10 percentage points (Llama-2-7b), while reducing memory usage by approximately 76\%. This indicates that in many scenarios, similar performance can be achieved with significantly lower GPU RAM usage.

\subsection{GPU Usage vs. Inference Time}
\label{sec: f1_vs_time}

\begin{figure}
\centering
\includegraphics[width=\columnwidth]{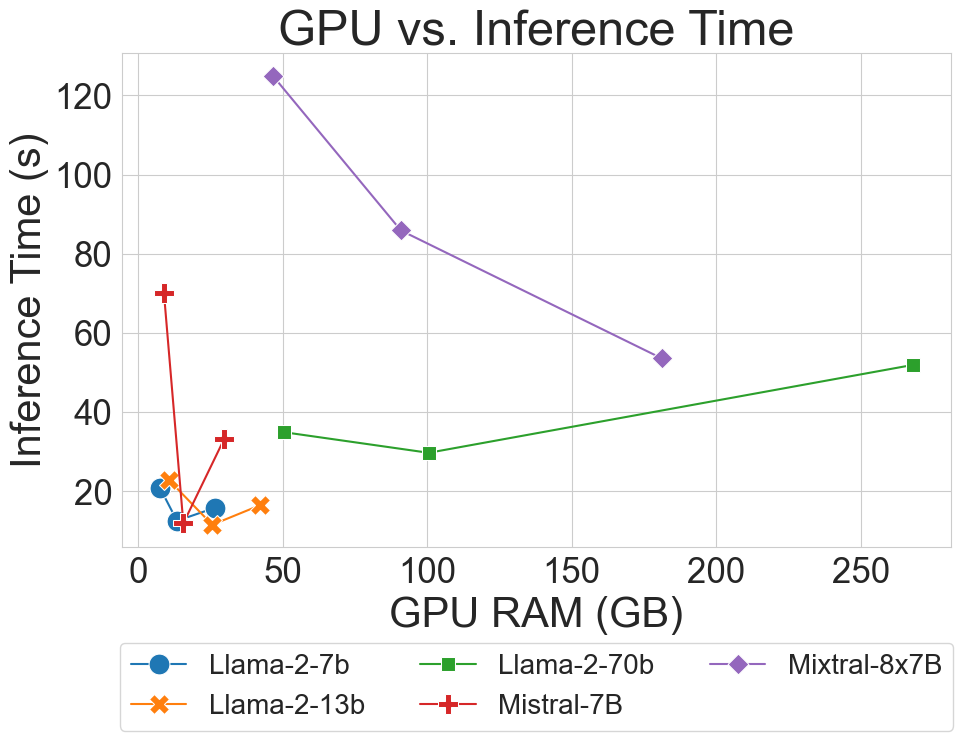}
\caption{GPU usage (GB) vs. inference time (seconds) per text instance.}
\label{fig:ram vs time}
\end{figure}

Table \ref{table:quantization performance} and Figure \ref{fig:ram vs time} illustrate the comparative analysis of GPU usage and inference time for each model configuration. Generally, as quantization increases, inference time also increases, except for Mixtral-8X7b. More aggressive quantization leads to longer generation times. In most cases, generating an instance with 8-bit quantization takes about twice as long as with 32-bit quantization. However, this increased time is offset by a roughly 76\% reduction in GPU RAM.

\subsection{Qualitative Analysis}
\label{sec: quatiltive analysis}

The comparative analysis (Appendix \ref{appendix:affective text generation}, Table \ref{tab:affective_text}) reveals a qualitative distinctions in the text produced by the LLMs. While both the 7B and 70B models demonstrate similar performance in raw emotion scoring from the classifier, a closer examination of the output text suggests that the 70B models yield superior and more coherent results. This is evident in the more nuanced and detailed responses to prompts about emotions such as anger. The 70B-8 and 70B-32 configurations, in particular, generate text that more effectively captures the complexity of the emotional scenarios, providing richer context and depth in the examples given. Similar results can be observed with Mixtral-8X7b. For Mistral-7b, quantization has minimal impact on the generated text, with consistency maintained across different quantization levels, as show in previous sections. This suggests that larger models may have a better capacity for generating text that not only aligns with the desired emotion but also delivers a clearer and more precise narrative, indicative of a deeper understanding and processing of affective content.

\section{Conclusion and Future Work}
\label{sec: conclusion}

In summary, we explore the effects of quantization on LLMs for affective text generation. Our findings indicate that reducing precision bits leads to a performance decrease of up to 10 pp in F$_1$ score for larger models. In contrast, smaller models have an increase of up to 10 pp. While this reduction in precision does indeed affect model performance in affective text generation, it offers a substantial benefit in terms of memory consumption, with a remarkable reduction of 76\%. In terms of text quality, larger models at lower quantization levels generally outperform smaller, higher-precision models while requiring similar memory. This trade-off between efficiency and efficacy is a critical consideration for deploying LLMs in resource-constrained environments where memory usage is a limiting factor.

Further research will explore the quantization trade-off in additional NLP tasks (e.g., prompt optimization), different languages, and quantization techniques (e.g., quantization-aware training).

\section*{Acknowledgements}
 This work has been supported by a CONACYT scholarship
 (2020-000009-01EXTF-00195) and by the German
 Research Council (DFG), project
 ``Computational Event Analysis based on Appraisal Theories for Emotion
 Analysis'' (CEAT, project number KL 2869/1-2).

\section{Ethical Considerations}
\label{sec:ethical_considerations}

When deploying large language models (LLMs) for affective text generation, ethical considerations must be addressed, particularly in the context of quantization and its impact on model performance and resource usage. The reduction in precision and corresponding decrease in F$_1$ score for larger models may lead to less accurate or coherent text generation, which could affect user experience and trust, especially in applications requiring high sensitivity, such as mental health support or personalized recommendations. Additionally, the increased inference time associated with quantization could lead to slower response times, potentially frustrating users and reducing the effectiveness of real-time applications.

Finally, the underlying risks associated with the base pre-trained language models (e.g., Llama-2, Mistral) must be considered. These models may have been trained on biased data, potentially leading to text that perpetuates stereotypes or marginalizes certain groups. It's important to note that such risks are not inherent to the quantization methodology but stem from the base models used.

\section{Limitations}
While our study provides valuable insights into the effects of quantization on large language models for affective text generation, several limitations should be acknowledged. Our evaluation is limited to specific models (Llama-2 and Mistral), and results may vary with other architectures and parameter sizes. The study uses a single emotion dataset, and larger, more diverse datasets could offer more comprehensive insights. Additionally, while we noted an increase in inference time, the study does not deeply investigate computational overhead, potential optimizations, or the impact of different GPU hardware.

\bibliography{custume.bib}

\newpage

\appendix

\section{Automatic Text Quality Evaluation}
\label{appendix:gpt3 QE}

To complement the GPT-4o automatic text quality evaluation in Section \ref{sub: text quality}, we conducted the same experimental setup with GPT-3.5 to align with \citet{chiang-lee-2023-closer, liu-etal-2023-g}. Table \ref{text quality} presents the results for the 15 experimental configurations, evaluating Coherence, Fluency, Grammar, Native Speaker Likeness, Human Likeness, and Plausibility. We used a 5-point Likert scale, ranging from 1 (disagree) to 5 (agree), to quantify each dimension of text quality. Overall, the results from GPT-3.5 align with those from GPT-4o, though with slightly lower scores across all models.

\begin{table}
\centering
\footnotesize
\setlength{\tabcolsep}{3pt}
\begin{tabularx}{\columnwidth}{Xrrrrrr}
\toprule
\textbf{Model} & \textbf{Qnt.} & \textbf{Flu} & \textbf{N. Spk.} & \textbf{Coh.} & \textbf{Pla.} & \textbf{Gra.} \\
\midrule

Llama-2-7b & 8 &  3.3  & 3.6 & 2.4 & 3.0 & 3.5 \\
Llama-2-7b & 16 & 3.2 & 3.5 &  2.5 & 2.7 & 3.5\\
Llama-2-7b & 32 & 3.2 & 3.5 &  2.2 & 2.4 & 3.5\\
\midrule
Llama-2-13b & 8 & 3.5 & 3.5 & 2.5 & 2.6 & 3.5 \\
Llama-2-13b & 16 & 3.1 & 3.6 & 2.3  & 2.5 & 3.5\\
Llama-2-13b & 32 & 3.3 & 3.7 & 2.4  & 2.7 & 3.5\\

\midrule
Llama-2-70b & 8 & 3.7 & 4.0 & 3.2 & 3.5 & 3.8\\
Llama-2-70b & 16 & 3.8 & 4.0 & 3.5  & 3.6 & 3.9\\
Llama-2-70b & 32 & 3.4 & 3.6 & 2.4  & 2.8 & 3.6\\

\midrule
Mistral-7b & 8 & 4.1 & 4.0 & 3.8  & 3.8 & 3.8 \\
Mistral-7b & 16 & 4.1 & 4.1 & 3.9 & 3.8 & 3.9\\
Mistral-7b & 32 & 4.1 & 4.1 &  4.0 &  3.7 & 3.9\\

\midrule
Mixtral-8X7b & 8 & 3.5 & 3.8 & 2.4 & 3.3 & 3.5\\
Mixtral-8X7b & 16 & 4.1 & 4.0 & 4.0 & 3.7 & 3.9\\
Mixtral-8X7b & 32 & 4.1 & 4.0 & 3.9 & 3.7 & 3.9\\

\bottomrule
\end{tabularx}
\caption{Text quality evaluation using a five-level Likert scale for fluency (Flu), native speaker (N. Spk), coherency (Coh.), plausibility (Pla.), and overall grammar (Gra.). A score of 1 indicates "not at all agree," and 5 indicates "extremely agree" (higher is better).}
\label{table:gpt 3 text quality}
\end{table}

\section{Affective Text Generation}
\label{appendix:affective text generation}

We create an \textit{Emotion-Prompt} that combines each seed prompt (e.g., "Write a text that expresses \emoclass.") with one of seven emotions: Joy, Fear, Anger, Sadness, Disgust, Shame, and Guilt. Here, \emoclass is replaced by each emotion category (see Table \ref{tab:emotion_prompts} for examples).

\begin{table}
\centering \footnotesize
\begin{tabularx}{\columnwidth}{lXr}
\toprule
\textbf{Emotion} & \textbf{Emotion-Prompt Example} \\
\midrule
Joy & Phrases that express \texttt{Joy} \\
Fear & Phrases that express \texttt{Fear} \\
Anger & Phrases that express \texttt{Anger} \\
Sadness & Phrases that express \texttt{Sadness} \\
Disgust & Phrases that express \texttt{Disgust} \\
Shame & Phrases that express \texttt{Shame} \\
Guilt & Phrases that express \texttt{Guilt} \\
\bottomrule
\end{tabularx}
\caption{Examples of Emotion-Prompts for each emotion category for the
  prompt ``\textit{Phrases that express \emoclass.}''.}
\label{tab:emotion_prompts}
\end{table}

We generate affective text from five LLMs using three different configurations and the \textit{Emotion-Prompt}. A beam search strategy was employ to produce the three most likely sentences, using a beam size of 30, a next-token temperature of 0.7, and a top-p (nucleus) sampling strategy set to 0.7. To ensure quality, we excluded sentences containing duplicate bigrams. Table \ref{tab:affective_text} shows a sample of the generated text from different configurations using the same \textit{Emotion-Prompt}.\\[5pt]

\onecolumn
\begingroup
\footnotesize
  \setlength{\tabcolsep}{2pt}
\renewcommand{\arraystretch}{1}
\noindent
\begin{tabularx}{\textwidth}{lX}
\toprule
\textbf{Model} & \textbf{Generated text} \\
\midrule
Llama-2-7B-8b & "Here are some examples of situations that can cause anger. How to write a descriptive essay about a place. A descriptive essay is a type of writing that uses sensory details to describe a place, person, object, or experience. The goal of a descriptive essay is to create a vivid image in the reader's mind and to convey the writer's feelings and perceptions about the subject." \\
Llama-2-7B-32b & ", but was unable to express it in a healthy way. A person may feel anger but be unable to express it in a healthy way due to various reasons. Here are some possible scenarios: 1. Fear of reprisal: The person may be afraid of the consequences of expressing their anger, such as losing their job, damaging their relationship, or facing social ridicule." \\
\midrule
Llama-2-13B-8b & 4. Describe a situation where a person felt fear.
5. Describe a situation where a person felt happiness.
6. Describe a situation where a person felt sadness.
7. Describe a situation where a person felt surprise.
8. Describe a situation where a person felt embarrassment.
9. Describe a situation where a person felt disgust.\\
Llama-2-13B-32b  & What was the situation? What did they do with the anger? Did it change the outcome of the situation?
5. Describe a time when you were happy. What happened? Why were you happy? Did this happiness change your life in some way?
6. Describe a time when you were sad. What happened? Why were you sad? Did this sadness change your life in some way?
7. Describe a time \\
\midrule
Llama-2-70B-8b & " * What was the trigger for the anger? * How did the person express their anger? * How did the situation unfold, and what was the outcome? * How did the person feel after the situation had resolved? * What did the person learn from the experience? Example: Situation: A colleague took credit for a project that Jane had worked on for months, and" \\
Llama-2-70B-32b & "You have been betrayed by someone you considered a friend. You trusted this person with your deepest secrets and they have gone behind your back and shared them with others. You feel angry and hurt by their actions." \\

\midrule

Mistral-7B-8b & and Question 2: How did the person express their anger?\\

Mistral-7B-32b &  and Question 2: How did the person express their anger? \\
\midrule
Mixtral-8x7B-8b &  and

(a) expressed it openly, (b) suppressed it.

Answer:

(a) A student in my class was constantly disturbing the teacher during her lecture by talking loudly with his friends. The teacher asked him to keep quiet several times but he did not pay any attention. Finally, she lost her temper and scolded him in front of the whole class. She told him that if he didn’t want to study, he should leave the class immediately. This made the student feel embarrassed and angry at the same time. However, he controlled his anger and apologized to the teacher for his behavior.

(b) My younger brother had borrowed my new bicycle without asking me and damaged its chain while riding it roughly on the road. When I came to know about this, I felt very angry as the bicycle was quite expensive and I had saved money from my pocket money to buy it. However, I suppressed my anger as I didn’t want to create a scene in front of my parents. I calmly explained to him how careless he had been and made him promise that he would never do such a thing again. I also made him pay for the repair of the bicycle from his own pocket money. \\
Mixtral-8x7B-32b &   and

a) The person was angry because they were not given the opportunity to speak in a meeting. They had prepared some important points that they wanted to share with the group, but the chairperson did not call on them. This made them feel frustrated and unheard, which led to feelings of anger.

b) A friend of mine was angry because they found out that their partner had been cheating on them. They felt betrayed and hurt by this revelation, which caused them to experience intense anger towards their partner.

c) I once felt angry when I was stuck in traffic for hours due to an accident on the highway. I had planned to arrive at my destination at a certain time, but the delay caused me to miss an important appointment. This made me feel frustrated and upset, leading to feelings of anger.\\ 
\bottomrule
\end{tabularx}
\captionof{table}{Affective text examples generated by five LLMs at two quantization levels (8 and 32 bits) using the \textit{Emotion-Prompt}: ``Describe a situation where a person felt anger''}
\label{tab:affective_text}
\endgroup

\end{document}